%% file: main.tex
\documentclass[sigconf]{acmart}

\usepackage{subfiles}
\usepackage[flushleft]{threeparttable}

\AtBeginDocument{%
  \providecommand\BibTeX{{%
    \normalfont B\kern-0.5em{\scshape i\kern-0.25em b}\kern-0.8em\TeX}}}

\renewcommand\footnotetextcopyrightpermission[1]{}
\begin{document}

\begin{titlepage}
\begin{center}
 
\textsc{\Large  FinBERT: Financial Sentiment Analysis with Pre-trained Language Models }
\bigskip
\\
\textsc{\large
submitted in partial fulfillment for the degree of master of science\\
\bigskip
Dogu Araci\\
12255068\\
\bigskip
master information studies\\
data science \\
faculty of science\\
university of amsterdam\\
\bigskip
2019-06-25
}
\end{center}
 
\vfill
% In case of an internal project, remove External Supervisor or if you had two internal supervisors, change the header into 
%  & First Supervisor & Second Supervisor  \\
\begin{center}
\begin{tabular}{|l||ll|}
\hline
 & \textbf{Internal  Supervisor} & \textbf{External   Supervisor}  \\   
 \hline
\textbf{Title, Name} & Dr Pengjie Ren& Dr Zulkuf Genc  \\
\textbf{Affiliation} &UvA, ILPS & Naspers Group\\ 
\textbf{Email} & p.ren@uva.nl&zulkuf.genc@naspers.com \\
\hline
\end{tabular}
\end{center}
%% If you have a third supervisor use this table instead
%\begin{center}
%\begin{tabular}{|l||lll|}
%\hline
% & \textbf{External   Supervisor} & \textbf{External   Supervisor} & \textbf{3$^{\mathrm{rd}}$ supervisor} \\
% \hline
%\textbf{Title, Name} & Dr Maarten Marx& & \\
%\textbf{Affiliation} &UvA, FNWI, IvI & & \\ 
%\textbf{Email} & maartenmarx@uva.nl& &  .\\
%\hline
%\end{tabular}
%\end{center}
\bigskip
% logos
\begin{center}
\mbox{\includegraphics[width=.2\paperwidth]{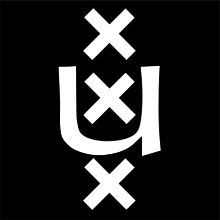} 
\includegraphics[width=.2\paperwidth]{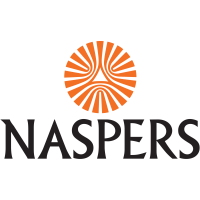} % replace by the logo of your internship company or remove
}
\end{center}
\end{titlepage}

\title{FinBERT: Financial Sentiment Analysis with Pre-trained Language Models}
\author{Dogu Tan Araci}
\email{dogu.araci@student.uva.nl}
\affiliation{%
  \institution{University of Amsterdam}
  \city{Amsterdam}
  \state{The Netherlands}
}

\begin{abstract}

\subfile{text/abstract.tex}

\end{abstract}


\maketitle

\section{Introduction}

\subfile{text/introduction.tex}

\subfile{text/lit_review.tex}

\subfile{text/method.tex}

\subfile{text/exp_setup.tex}

\subfile{text/exp_results.tex}

\subfile{text/exp_analysis.tex}

\subfile{text/conclusion.tex}

\bibliographystyle{ACM-Reference-Format}

\bibliography{main}


\end{document}

%% file: text/abstract.tex
    Financial sentiment analysis is a challenging task due to the specialized language and lack of labeled data in that domain. General-purpose models are not effective enough because of specialized language used in financial context. We hypothesize that pre-trained language models can help with this problem because they require fewer labeled examples and they can be further trained on domain-specific corpora. We introduce FinBERT, a language model based on BERT, to tackle NLP tasks in financial domain. Our results show improvement in every measured metric on current state-of-the-art results for two financial sentiment analysis datasets. We find that even with a smaller training set and fine-tuning only a part of the model, FinBERT outperforms state-of-the-art machine learning methods.

%% file: text/introduction.tex
Prices in an open market reflects all of the available information regarding assets exchanged in an economy \cite{Malkiel2003}. When new information becomes available, all actors in the economy update their positions and prices adjust accordingly, which makes beating the markets consistently impossible. However, the definition of "new information" might change as new information retrieval technologies become available and early-adoption of such technologies might provide an advantage in the short-term. 

Analysis of financial texts, be it news, analyst reports or official company announcements is a possible source of new information. With unprecedented amount of such text being created every day, manually analyzing these and deriving actionable insights from them is too big of a task for any single entity. Hence, automated sentiment or polarity analysis of texts produced by financial actors using natural language processing (NLP) methods has gained popularity during the last decade \cite{Guo2016}.

The principal research interest for this thesis is the polarity analysis, which is classifying text as positive, negative or neutral, in a specific domain. It requires to address two challenges: 1) The most sophisticated classification methods that make use of neural nets require vast amounts of labeled data and labeling financial text snippets requires costly expertise. 2) The sentiment analysis models trained on general corpora are not suited to the task, because financial texts have a specialized language with unique vocabulary and have a tendency to use vague expressions instead of easily-identified negative/positive words. 

Using carefully crafted financial sentiment lexicons such as Loughran and McDonald (2011) \cite{Loughran2011} may seem a solution because they incorporate existing financial knowledge into textual analysis. However, they are based on "word counting" methods, which come short in analyzing deeper semantic meaning of a given text.

NLP transfer learning methods look like a promising solution to both of the challenges mentioned above, and are the focus of this thesis. The core idea behind these models is that by training language models on very large corpora and then initializing down-stream models with the weights learned from the language modeling task, a much better performance can be achieved. The initialized layers can range from the single word embedding layer \cite{Peters2018} to the whole model \cite{Howard2018}. This approach should, in theory, be an answer to the scarcity of labeled data problem. Language models don't require any labels, since the task is predicting the next word. They can learn how to represent the semantic information. That leaves the fine-tuning on labeled data only the task of learning how to use this semantic information to predict the labels.

One particular component of the transfer learning methods is the ability to further pre-train the language models on domain specific unlabeled corpus. Thus, the model can learn the semantic relations in the text of the target domain, which is likely to have a different distribution than a general corpus. This approach is especially promising for a niche domain like finance, since the language and vocabulary used is dramatically different than a general one.   

The goal of this thesis is to test these hypothesized advantages of using and fine-tuning pre-trained language models for financial domain. For that, sentiment of a sentence from a financial news article towards the financial actor depicted in the sentence will be tried to be predicted, using the Financial PhraseBank created by Malo et al. (2014) \cite{Malo2014} and FiQA Task 1 sentiment scoring dataset \cite{Maia2018}. 

The main contributions of this thesis are the following:
\begin{itemize}
    \item We introduce FinBERT, which is a language model based on BERT for financial NLP tasks. We evaluate FinBERT on two financial sentiment analysis datasets. 
    \item We achieve the state-of-the-art on FiQA sentiment scoring and Financial PhraseBank.
    \item We implement two other pre-trained language models, ULMFit and ELMo for financial sentiment analysis and compare these with FinBERT.  
    \item We conduct experiments to investigate several aspects of the model, including: effects of further pre-training on financial corpus, training strategies to prevent catastrophic forgetting and fine-tuning only a small subset of model layers for decreasing training time without a significant drop in performance. 
\end{itemize}

The rest of the thesis is structured as follows: First, relevant literature in both financial polarity analysis and pre-trained language models are discussed (Section \ref{sec:related_lit}). Then, the evaluated models are described (Section \ref{sec:method}). This is followed by the description of the experimental setup being used (Section \ref{exp_setup}). In Section \ref{exp_results}, we present the experimental results on the financial sentiment datasets. Then we further analyze FinBERT from different perspectives in Section \ref{exp_analysis}. Finally, we conclude with Section \ref{conclusion}.

%% file: text/lit_review.tex
\section{Related Literature} \label{sec:related_lit}
This section describes previous research conducted on sentiment analysis in finance (\ref{sentiment analysis in finance}) and text classification using pre-trained language models (\ref{text classification}).

\subsection{Sentiment analysis in finance} \label{sentiment analysis in finance}
Sentiment analysis is the task of extracting sentiments or opinions of people from written language \cite{Liu2012}. We can divide the recent efforts into two groups: 1) Machine learning methods with features extracted from text with "word counting" \cite{Agarwal2016,Whitelaw2005,conf/icwsm/MartineauF09,Tripathy2016}, 2) Deep learning methods, where text is represented by a sequence of embeddings \cite{Severyn2015,Araque2017,Zhang2018}. The former suffers from inability to represent the semantic information that results from a particular sequence of words, while the latter is often deemed as too "data-hungry" as it learns a much higher number of parameters \cite{2018arXiv180100631M}.  

Financial sentiment analysis differs from general sentiment analysis not only in domain, but also the purpose. The purpose behind financial sentiment analysis is usually guessing how the markets will react with the information presented in the text \cite{Li2014}. Loughran and McDonald (2016) presents a thorough survey of recent works on financial text analysis utilizing machine learning with "bag-of-words" approach or lexicon-based methods \cite{Loughran2016}. For example, in Loughran and McDonald (2011), they create a dictionary of financial terms with assigned values such as "positive" or "uncertain" and  measure the tone of a documents by counting words with a specific dictionary value  \cite{Loughran2011}. Another example is Pagolu et al. (2016), where n-grams from tweets with financial information are fed into supervised machine learning algorithms to detect the sentiment regarding the financial entity mentioned.

On of the first papers that used deep learning methods for textual financial polarity analysis was Kraus and Feuerriegel (2017) \cite{Kraus2017}. They apply an LSTM neural network to ad-hoc company announcements to predict stock-market movements and show that method to be more accurate than traditional machine learning approaches. They find pre-training their model on a larger corpus to improve the result, however their pre-training is done on a labeled dataset, which is a more limiting approach then ours, as we pre-train a language model as an unsupervised task.

There are several other works that employ various types of neural architectures for financial sentiment analysis. Sohangir et al. (2018) \cite{Sohangir2018} apply several generic neural network architectures to a StockTwits dataset, finding CNN as the best performing neural network architecture. Lutz et al. 2018 \cite{Lutz2018} take the approach of using \textit{doc2vec} to generate sentence embeddings in a particular company ad-hoc announcement and utilize multi-instance learning to predict stock market outcomes. Maia et al. (2018) \cite{Maia20182} use a combination of text simplification and LSTM network to classify a set of sentences from financial news according to their sentiment and achieve state-of-the-art results for the Financial PhraseBank, which is used in thesis as well.

Due to lack of large labeled financial datasets, it is difficult to utilize neural networks to their full potential for sentiment analysis. Even when their first (word embedding) layers are initialized with pre-trained values, the rest of the model still needs to learn complex relations with relatively small amount of labeled data. A more promising solution could be initializing almost the entire model with pre-trained values and fine-tuning those values with respect to the classification task.

\subsection{Text classification using pre-trained language models} \label{text classification}

Language modeling is the task of predicting the next word in a given piece of text. One of the most important recent developments in natural language processing is the realization that a model trained for language modeling can be successfully fine-tuned for most down-stream NLP tasks with small modifications. These models are usually trained on very large corpora, and then with addition of suitable task-specific layers fine-tuned on the target dataset \cite{Kant2018}. Text classification, which is the focus of this thesis, is one of the obvious use-cases for this approach.

ELMo (Embeddings from Language Models) \cite{Peters2018} was one of the first successful applications of this approach. With ELMo, a deep bidirectional language model is pre-trained on a large corpus. For each word, hidden states of this model is used to compute a contextualized representation. Using the pre-trained weights of ELMo, contextualized word embeddings can be calculated for any piece of text. Initializing embeddings for down-stream tasks with those were shown to improve performance on most tasks compared to static word embeddings such as word2vec or GloVe. For text classification tasks like SST-5, it achieved state-of-the-art performance when used together with a bi-attentive classification network \cite{McCann2017}. 

Although ELMo makes use of pre-trained language models for contextualizing representations, still the information extracted using a language model is present only in the first layer of any model using it. ULMFit (Universal Language Model Fine-tuning) \cite{Howard2018} was the first paper to achieve true transfer learning for NLP, as using novel techniques such as discriminative fine-tuning, slanted triangular learning rates and gradual unfreezing. They were able to efficiently fine-tune a whole pre-trained language model for text classification. They also introduced further pre-training of the language model on a domain-specific corpus, assuming target task data comes from a different distribution than the general corpus the initial model was trained on. 

ULMFit's main idea of efficiently fine-tuning a pre-trained a language model for down-stream tasks was brought to another level with Bidirectional Encoder Representations from Transformers (BERT) \cite{Devlin2018}, which is also the main focus of this paper. BERT has two important differences from what came before: 1) It defines the task of language modeling as predicting randomly masked tokens in a sequence rather than the next token, in addition to a task of classifying two sentences as following each other or not. 2) It is a very big network trained on an unprecedentedly large corpus. These two factors enabled in to achieve state-of-the-art results in multiple NLP tasks such as, natural language inference or question answering.

The specifics of fine-tuning BERT for text classification has not been researched thoroughly. One such recent work is Sun et al. (2019) \cite{Sun2019}. They conduct a series of experiments regarding different configurations of BERT for text classification. Some of their results will be referenced throughout the rest of the thesis, for the configuration of our model.

%% file: text/method.tex
\section{Method} \label{sec:method}

In this section, we will present our BERT implementation for financial domain named as FinBERT, after giving a brief background on relevant neural architectures.

\subsection{Preliminaries} \label{neural baselines}

\subsubsection{LSTM}
Long short-term memory (LSTM) is a type of recurrent neural network that allows long-term dependencies in a sequence to persist in the network by using "forget" and "update" gates. It is one of the primary architectures for modeling any sequential data generation process, from stock prices to natural language. 
Since a text is a sequence of tokens, the first choice for any LSTM natural language processing model is determining how to initially represent a single token. Using pre-trained weights for initial token representation is the common practice. One such pre-training algorithm is GLoVe (Global Vectors for Word Representation) \cite{pennington-etal-2014-glove}. GLoVr is a model for calculating word representations with the unsupervised task of training a log-bilinear regression model on a word-word co-occurance matrix from a large corpus. It is an effective model for representing words in a vector space, however it doesn't contextualize these representations with respect to the sequence they are actually used in\footnote{The pre-trained weights for GLoVE can be found here: https://nlp.stanford.edu/projects/glove/}.

\subsubsection{ELMo} \label{basic lstm models}
 ELMo embeddings \cite{Peters2018} are contextualized word representations in the sense that the surrounding words influence the representation of the word. In the center of ELMo, there is a bidirectional language model with multiple LSTM layers. The goal of a language model is to learn the probability distribution over sequences of tokens in a given vocabulary. ELMo models the probability of a token given the previous (and separately following) tokens in the sequence. Then the model also learns how to weight different representations from different LSTM layers in order to calculate one contextualized vector per token. Once the contextualized representations are extracted, these can be used to initialize any down-stream NLP task\footnote{The pre-trained ELMo models can be found here: https://allennlp.org/elmo}.
 
 \subsubsection{ULMFit} \label{ulmfit}
 ULMFit is a transfer learning model for down-stream NLP tasks, that make use of language model pre-training \cite{Howard2018}. Unlike ELMo, with ULMFit, the whole language model is fine-tuned together with the task-specific layers. The underlying language model used in ULMFit is AWD-LSTM, which uses sophisticated dropout tuning strategies to better regularize its LSTM model \cite{DBLP:journals/corr/abs-1708-02182}. For classification using ULMFit two linear layers are added to the pre-trained AWD-LSTM, first of which takes the pooled last hidden states as input. 
 
ULMFit comes with novel training strategies for further pre-training the language model on domain-specific corpus and fine-tuning on the down-stream task. We implement these strategies with FinBERT as explained in section \ref{finbert}.
 
 \subsubsection{Transformer}
 The Transformer is an attention-based architecture for modeling sequential information, that is an alternative to recurrent neural networks \cite{Vaswani2017}. It was proposed as a sequence-to-sequence model, therefore including encoder and decoder mechanisms. Here, we will focus only on the encoder part (though decoder is quite similar). The encoder consists of multiple identical Transformer layers. Each layer has a multi-headed self-attention layer and a fully connected feed-forward network. For one self-attention layer, three mappings from embeddings (key, query and value) are learned. Using each token's key and all tokens' query vectors, a similarity score is calculated with dot product. These scores are used to weight the value vectors to arrive at the new representation of the token. With the multi-headed self-attention, these layers are concatenated together, so that the sequence can be evaluated from varying "perspectives". Then the resulted vectors go through fully connected networks with shared parameters. 
 
 As it was argued by Vaswani 2017 \cite{Vaswani2017}, Transformer architecture has several advantages over the RNN-based approaches. Because of RNNs' sequential nature, they are much harder to parallelize on GPUs and too many steps between far away elements in a sequence make it hard for information to persist.
 
\subsubsection{BERT}
 BERT \cite{Devlin2018} is in essence a language model that consists of a set of Transformer encoders stacked on top of each other. However it defines the language modeling task differently from ELMo and AWD-LSTM. Instead of predicting the next word given previous ones, BERT "masks" a randomly selected 15\% of all tokens. With a softmax layer over vocabulary on top of the last encoder layer the masked tokens are predicted. A second task BERT is trained on is "next sentence prediction". Given two sentences, the model predicts whether or not these two actually follow each other. 
 
 The input sequence is represented with token and position embeddings. Two tokens denoted by \textsc{[CLS]} and \textsc{[SEP]} are added to the beginning and end of the sequence respectively. For all classification tasks, including the next sentence prediction, \textsc{[CLS]} token is used. 
 
 BERT has two versions: BERT-base, with 12 encoder layers, hidden size of 768, 12 multi-head attention heads and 110M parameters in total and BERT-large, with 24 encoder layers, hidden size of 1024, 16 multi-head attention heads and 340M parameters. Both of these models have been trained on BookCorpus \cite{Zhu2015} and English Wikipedia, which have in total more than 3,500M words \footnote{The pre-trained weights are made public by creators of BERT. The code and weights can be found here: https://github.com/google-research/bert}. 
  
\subsection{BERT for financial domain: FinBERT} \label{finbert}
In this subsection we will describe our implementation of BERT: 1) how further pre-training on domain corpus is done, 2-3) how we implemented BERT for classification and regression tasks, 4) training strategies we used during fine-tuning to prevent catastrophic forgetting.
\subsubsection{Further pre-training} \label{further pre-training}
Howard and Ruder (2018) \cite{Howard2018} shows that futher pre-training a language model on a target domain corpus improves the eventual classification performance. For BERT, there is not decisive research showing that would be the case as well. Regardless, we implement further pre-training in order to observe if such adaptation is going to be beneficial for financial domain.

For further pre-training, we experiment with two approaches. The first is pre-training the model on a relatively large corpus from the target domain. For that, we further pre-train a BERT language model on a financial corpus (details of the corpus can be found on section \ref{trc2}). The second approach is pre-training the model only on the sentences from the training classification dataset. Although the second corpus is much smaller, using data from the direct target might provide better target domain adaptation.

\subsubsection{FinBERT for text classification} \label{classification}
\begin{figure*}[h] 
  \centering
  \includegraphics[width=\textwidth]{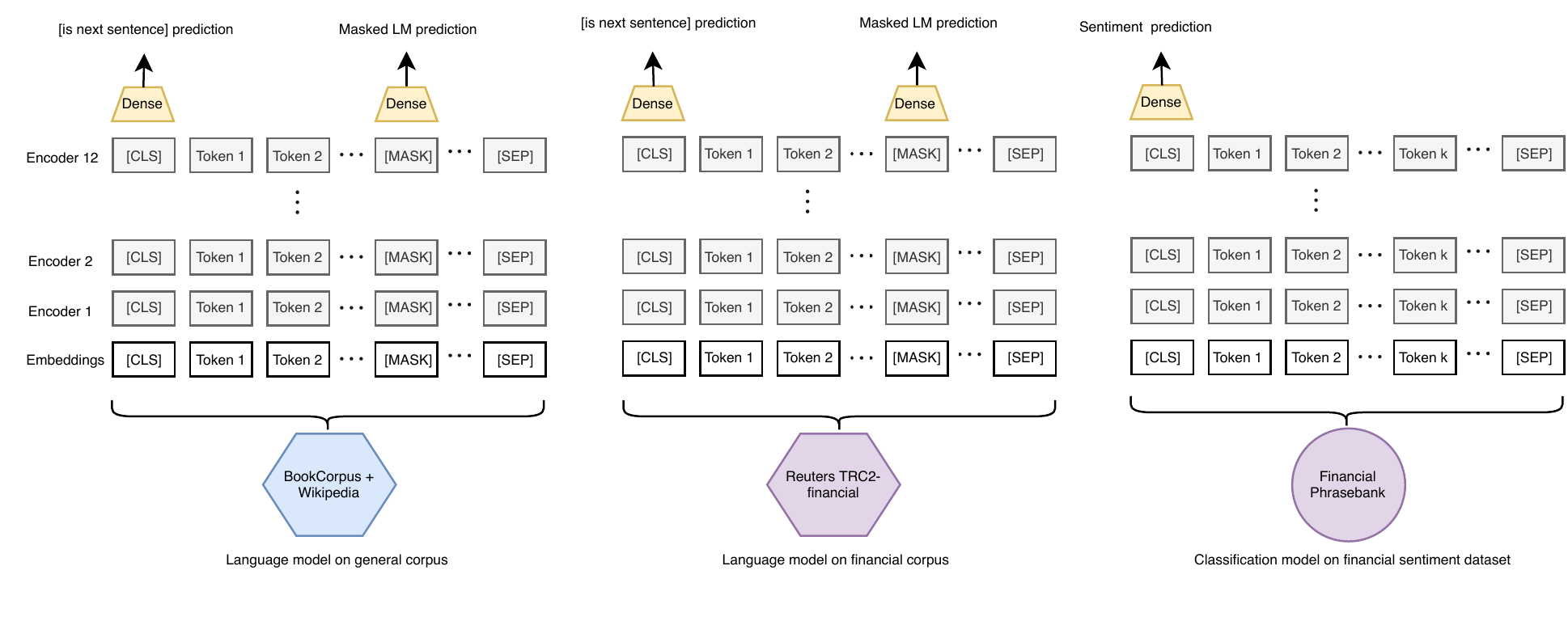}
  \caption{Overview of pre-training, further pre-training and classification fine-tuning}
  \label{fig:model overview}
  \Description{The 1907 Franklin Model D roadster.}
\end{figure*}

Sentiment classification is conducted by adding a dense layer after the last hidden state of the \textsc{[CLS]} token. This is the recommended practice for using BERT for any classification task \cite{Devlin2018}. Then, the classifier network is trained on the labeled sentiment dataset. An overview of all the steps involved in the procedure is presented on figure \ref{fig:model overview}. 

\subsubsection{FinBERT for regression}
While the focus of this paper is classification, we also implement regression with almost the same architecture on a different dataset with continuous targets. The only difference is that the loss function being used is mean squared error instead of the cross entropy loss. 

\subsubsection{Training strategies to prevent catastrophic forgetting} 
As it was pointed out by Howard and Ruder (2018) \cite{Howard2018}, catastrophic forgetting is a significant danger with this fine-tuning approach. Because the fine-tuning procedure can quickly cause model to "forget" the information from language modeling task as it tries to adapt to the new task. In order to deal with this phenomenon, we apply three techniques as it was proposed by Howard and Ruder (2018): slanted triangular learning rates, discriminative fine-tuning and gradual unfreezing. 

Slanted triangular learning rate applies a learning rate schedule in the shape of a slanted triangular, that is, learning rate first linearly increases up to some point and after that point linearly decreases. 

Discriminative fine-tuning is using lower learning rates for lower layers on the network. Assume our learning rate at layer $l$ is $\alpha$. Then for discrimination rate of $\theta$ we calculate the learning rate for layer $l-1$ as $\alpha_{l-1} = \theta \alpha_l$. The assumption behind this method is that the lower layers represent the deep-level language information, while the upper ones include information for actual classification task. Therefore we fine-tune them differently. 

With gradual freezing, we start training with all layers but the classifier layer as frozen. During training we gradually unfreeze all of the layers starting from the highest one, so that the lower level features become the least fine-tuned ones. Hence, during the initial stages of training it is prevented for model to "forget" low-level language information that it learned from pre-training.

%% file: text/exp_setup.tex
\section{Experimental Setup} \label{exp_setup}

\subsection{Research Questions} \label{research_questions}
We aim to answer the following research questions:
\begin{itemize}
    \item[(RQ1)] What is the performance of FinBERT in short sentence classification compared with the other transfer learning methods like ELMo and ULMFit?
    \item[(RQ2)] How does FinBERT compare to the state-of-the-art in financial sentiment analysis with targets discrete or continuous? 
    \item[(RQ3)] How does futher pre-training BERT on financial domain, or target corpus, affect the classification performance?
    \item[(RQ4)] What are the effects of training strategies like slanted triangular learning rates, discriminative fine-tuning and gradual unfreezing on classification performance? Do they prevent catastrophic forgetting? 
    \item[(RQ5)] Which encoder layer performs best (or worse) for sentence classification?
    \item[(RQ6)] How much fine-tuning is enough? That is, after pre-training, how many layers should be fine-tuned to achieve comparable performance to fine-tuning the whole model?
\end{itemize}

\subsection{Datasets} \label{dataset}

\subsubsection{TRC2-financial} \label{trc2}
In order to further pre-train BERT, we use a financial corpus we call TRC2-financial. It is a subset of Reuters' TRC2\footnote{The corpus can be obtained for research purposes by applying here: https://trec.nist.gov/data/reuters/reuters.html}, which consists of 1.8M news articles that were published by Reuters between 2008 and 2010. We filter for some financial keywords in order to make corpus more relevant and in limits with the compute power available. The resulting corpus, TRC2-financial, includes 46,143 documents with more than 29M words and nearly 400K sentences.

\subsubsection{Financial PhraseBank}
The main sentiment analysis dataset used in this paper is Financial PhraseBank\footnote{The dataset can be found here: https://www.researchgate.net/publication/251231364 \_FinancialPhraseBank-v10} from Malo et al. 2014 \cite{Malo2014}. Financial Phrasebank consists of 4845 english sentences selected randomly from financial news found on LexisNexis database. These sentences then were annotated by 16 people with background in finance and business. The annotators were asked to give labels according to how they think the information in the sentence might affect the mentioned company stock price. The dataset also includes information regarding the agreement levels on sentences among annotators. The distribution of agreement levels and sentiment labels can be seen on table \ref{tab:phrasebank}. We set aside 20\% of all sentences as test and 20\% of the remaining as validation set. In the end, our train set includes 3101 examples. For some of the experiments, we also make use of 10-fold cross validation.

\subfile{tables/phrasebank.tex}

\subsubsection{FiQA Sentiment}
FiQA \cite{Maia2018} is a dataset that was created for WWW '18 conference financial opinion mining and question answering challenge\footnote{Data can be found here: https://sites.google.com/view/fiqa/home}. We use the data for Task 1, which includes 1,174 financial news headlines and tweets with their corresponding sentiment score. Unlike Financial Phrasebank, the targets for this datasets are continuous ranging between $[-1,1]$ with $1$ being the most positive. Each example also has information regarding which financial entity is targeted in the sentence. We do 10-fold cross validation for evaluation of the model for this dataset.

\subsection{Baseline Methods} \label{baseline_methods}
For contrastive experiments, we consider baselines with three different methods: LSTM classifier with GLoVe embeddings, LSTM classifier with ELMo embeddings and ULMFit classifier. It should be noted that these baseline methods are not experimented with as thoroughly as we did with BERT. Therefore the results should not be interpreted as definitive conclusions of one method being better. 

\subsubsection{LSTM classifiers}
We implement two classifiers using bidirectional LSTM models. In both of them, a hidden size of 128 is used, with the last hidden state size being 256 due to bidirectionality. A fully connected feed-forward layer maps the last hidden state to a vector of three, representing likelihood of three labels. The difference between two models is that one uses GLoVe embeddings, while the other uses ELMo embeddings. A dropout probability of 0.3 and a learning rate of 3e-5 is used in both models. We train them until there is no improvement in validation loss for 10 epochs.

\subsubsection{ULMFit}
As it was explained in section \ref{ulmfit}, classification with ULMFit consists of three steps. The first step of pre-training a language model is already done and the pre-trained weights are released by Howard and Ruder (2018). We first further pre-train AWD-LSTM language model on TRC2-financial corpus for 3 epochs. After that, we fine-tune the model for classification on Financial PhraseBank dataset, by adding a fully-connected layer to the output of pre-trained language model. 

\subsection{Evaluation Metrics} \label{evaluation_metrics}
For evaluation of classification models, we use three metrics: Accuracy, cross entropy loss and macro F1 average. We weight cross entropy loss with square root of inverse frequency rate. For example if a label constitutes 25\% of the all examples, we weight the loss attributed to that label by 2. Macro F1 average calculates F1 scores for each of the classes and then takes the average of them. Since our data, Financial PhraseBank suffers from label imbalance (almost 60\% of all sentences are neutral), this gives another good measure of the classification performance. For evaluation of regression model, we report mean squared error and $R^2$, as these are both standard and also reported by the state-of-the-art papers for FiQA dataset.

\subsection{Implementation Details} \label{implementation_details}
For our implementation BERT, we use a dropout probability of $p=0.1$, warm-up proportion of $0.2$, maximum sequence length of $64$ tokens, a learning rate of $2e-5$ and a mini-batch size of $64$. We train the model for 6 epochs, evaluate on the validation set and choose the best one. For discriminative fine-tuning we set the discrimination rate as 0.85. We start training with only the classification layer unfrozen, after each third of a training epoch we unfreeze the next layer. An Amazon p2.xlarge EC2 instance with one NVIDIA K80 GPU, 4 vCPUs and 64 GiB of host memory is used to train the models.

%% file: tables/phrasebank.tex
\begin{table}
  \caption{Distribtution of sentiment labels and agreement levels in Financial PhraseBank}
  \label{tab:phrasebank}
  \begin{tabular}{lcccc}
    \toprule
    Agreement level  &  Positive &  Negative &  Neutral & Count\\
    \midrule
    100\% &         \%25.2 & \%13.4 & \%61.4 & 2262 \\
    75\% - 99\% &   \%26.6 & \%9.8 & \%63.6 & 1191 \\
    66\% - 74\% &   \%36.7 & \%12.3 & \%50.9 & 765 \\
    50\% - 65\% &   \%31.1 & \%14.4 & \%54.5 & 627 \\
    \midrule
    All &           \%28.1 & \%12.4 & \%59.4 & 4845 \\
  \bottomrule
\end{tabular}
\end{table}

%% file: text/exp_results.tex
\section{Experimental Results (RQ1 \& RQ2)} \label{exp_results}

The results of FinBERT, the baseline methods and state-of-the-art on Financial PhraseBank dataset classification task can be seen on table \ref{tab:results}. We present the result on both the whole dataset and subset with 100\% annotator agreement.

\subfile{tables/results.tex}

For all of the measured metrics, FinBERT performs clearly the best among both the methods we implemented ourselves (LSTM and ULMFit) and the models reported by other papers (LPS \cite{Malo2014}, HSC \cite{Krishnamoorthy2018}, FinSSLX \cite{Maia20182}). LSTM classifier with no language model information performs the worst. In terms of accuracy, it is close to LPS and HSC, (even better than LPS for examples with full agreement), however it produces a low F1-score. That is due to it performing much better in neutral class. LSTM classifier with ELMo embeddings improves upon LSTM with static embeddings in all of the measured metrics. It still suffers from low average F1-score due to poor performance in less represented labels. But it's performance is comparable with LPS and HSC, besting them in accuracy. So contextualized word embeddings produce close performance to machine learning based methods for dataset of this size. 

ULMFit significantly improves on all of the metrics and it doesn't suffer from model performing much better in some classes than the others. It also handily beats the machine learning based models LPS and HSC. This shows the effectiveness of language model pre-training. AWD-LSTM is a very large model and it would be expected to suffer from over-fitting with this small of a dataset. But due to language model pre-training and effective training strategies, it is able to overcome small data problem. ULMFit also outperforms FinSSLX, which has a text simplification step as well as pre-training of word embeddings on a large financial corpus with sentiment labels. 

\begin{figure}[h] 
  \centering
  \includegraphics[width=\linewidth]{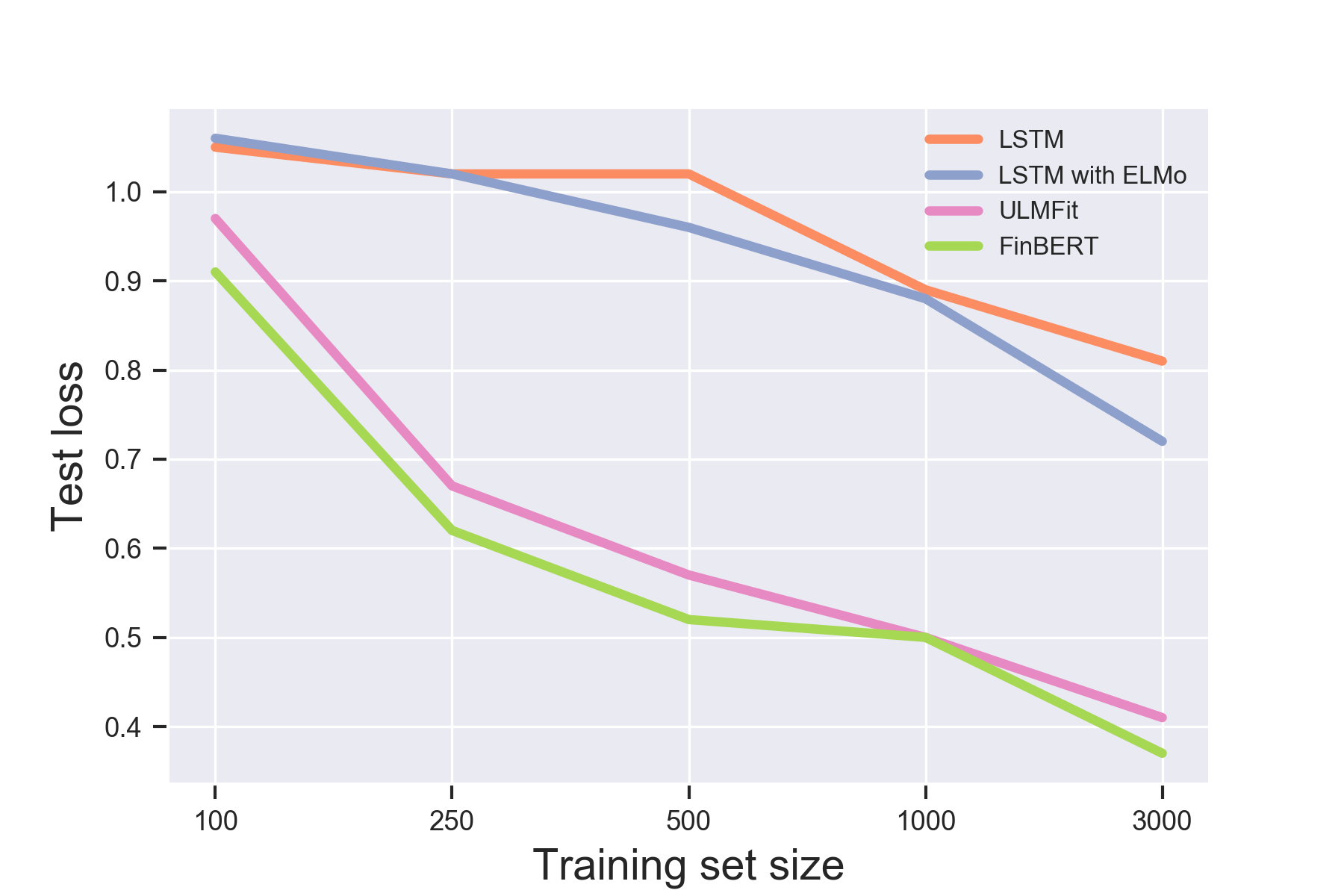}
  \caption{Test loss different training set sizes}
  \label{fig:loss}
  \Description{The 1907 Franklin Model D roadster.}
\end{figure}

FinBERT outperforms ULMFit, and consequently all of the other methods in all metrics. In order to measure the performance of the models on different sizes of labeled training datasets, we ran LSTM classifiers, ULMFit and FinBERT on 5 different configurations. The result can be seen on figure \ref{fig:loss}, where the cross entropy losses on test set for each model are drawn. 100 training examples is too low for all of the models. However, once the training size becomes 250, ULMFit and FinBERT starts to successfully differentiate between labels, with an accuracy as high as 80\% for FinBERT. All of the methods consistently get better with more data, but ULMFit and FinBERT does better with 250 examples than LSTM classifiers do with the whole dataset. This shows the effectiveness of language model pre-training.

The results for FiQA sentiment dataset, are presented on table \ref{tab:regression}. Our model outperforms state-of-the-art models for both MSE and $R^2$. It should be noted that the test set these two papers \cite{Yang2018} \cite{Piao2018} use is the official FiQA Task 1 test set. Since we don't have access to that we report the results on 10-Fold cross validation. There is no indication on \cite{Maia2018} that the train and test sets they publish come from different distributions and our model can be interpreted to be at disadvantage since we need to set aside a subset of training set as test set, while state-of-the-art papers can use the complete training set.

\subfile{tables/regression.tex}

%% file: tables/results.tex
\begin{table*}
\begin{threeparttable}
    \caption{Experimental Results on the Financial PhraseBank dataset}
  \label{tab:results}
  \begin{tabular}{lccccccc}
    \toprule
      &  \multicolumn{3}{c}{All data} & \multicolumn{3}{c}{Data with 100\% agreement}\\
      \cmidrule(lr){2-4}
      \cmidrule(lr){5-7}
      %\cmidrule{lr}{5-7}
      Model & Loss & Accuracy & F1 Score & Loss & Accuracy & F1 Score \\
    \midrule
    LSTM &                  0.81 & 0.71 & 0.64 & 0.57 & 0.81 & 0.74 \\
    LSTM with ELMo &        0.72 & 0.75 & 0.7 & 0.50 & 0.84 & 0.77 \\
    ULMFit &                0.41 & 0.83 & 0.79 & 0.20 & 0.93 & 0.91 \\
    \midrule
    LPS &                   - & 0.71 & 0.71 & - & 0.79 & 0.80 \\
    HSC &                   -  & 0.71 & 0.76 & - & 0.83 & 0.86 \\
    FinSSLX &                   -  & - & - & - & 0.91 & 0.88 \\
    \midrule
    FinBERT &               \textbf{0.37} & \textbf{0.86} & \textbf{0.84} & \textbf{0.13} & \textbf{0.97} & \textbf{0.95} \\
  \bottomrule
\end{tabular}
    \begin{tablenotes}
        \small
        \item \textbf{Bold face} indicates best result in the corresponding metric. LPS \cite{Malo2014}, HSC \cite{Krishnamoorthy2018} and FinSSLX \cite{Maia2018} results are taken from their respective papers. For LPS and HSC, overall accuracy is not reported on the papers. We calculated them using recall scores reported for different classes. For the models implemented by us, we report 10-fold cross validation results. 
    \end{tablenotes}
\end{threeparttable}
\end{table*}

%% file: tables/regression.tex
\begin{table}
\begin{threeparttable}
  \caption{Experimental Results on FiQA Sentiment Dataset}
  \label{tab:regression}
  \begin{tabular}{lcc}
    \toprule
    Model  &  MSE \hspace{6mm}&  $R^2$ \\
    \midrule
    Yang et. al. (2018) \hspace{20mm} &      0.08 & 0.40 \\
    Piao and Breslin (2018)  & 0.09 & 0.41 \\   
    \midrule
    FinBERT &      \textbf{0.07} & \textbf{0.55} \\
  \bottomrule
\end{tabular}
\begin{tablenotes}
\small 
\item \textbf{Bold face} indicated best result in corresponding metric. Yang et. al. (2018) \cite{Yang2018} and Piao and Breslin (2018) \cite{Piao2018} report results on the official test set. Since we don't have access to that set our MSE, and $R^2$ are calculated with 10-Fold cross validation. 
\end{tablenotes}
\end{threeparttable}
\end{table}

%% file: text/exp_analysis.tex
\section{Experimental Analysis} \label{exp_analysis}

\subsection{Effects of further pre-training (RQ3)}
We first measure the effect of further pre-training on the performance of the classifier. We  compare three models: 1) No further pre-training (denoted by Vanilla BERT), 2) Further pre-training on classification training set (denoted by FinBERT-task), 3) Further pre-training on domain corpus, TRC2-financial (denoted by FinBERT-domain). Models are evaluated with loss, accuracy and macro average F1 scores on the test dataset. The results can be seen on table \ref{tab:pretraining}.

\subfile{tables/finetune_results.tex}

The classifier that were further pre-trained on financial domain corpus performs best among the three, though the difference is not very high. There might be four reasons behind this result: 1) The corpus might have a different distribution than the task set, 2) BERT classifiers might not improve significantly with further pre-training, 3) Short sentence classification might not benefit significantly from further pre-training, 4) Performance is already so good, that there is not much room for improvement. We think that the last explanation is the likeliest, because for the subset of Financial Phrasebank that all of the annotators agree on the result, accuracy of Vanilla BERT is already 0.96. The performance on the other agreement levels should be lower, as even the humans can't agree fully on them. More experiments with another financial labeled dataset is necessary to conclude that effect of further pre-training on domain corpus is not significant.

\subsection{Catastrophic forgetting (RQ4)}
\begin{figure}[h] 
  \centering
  \includegraphics[width=\linewidth]{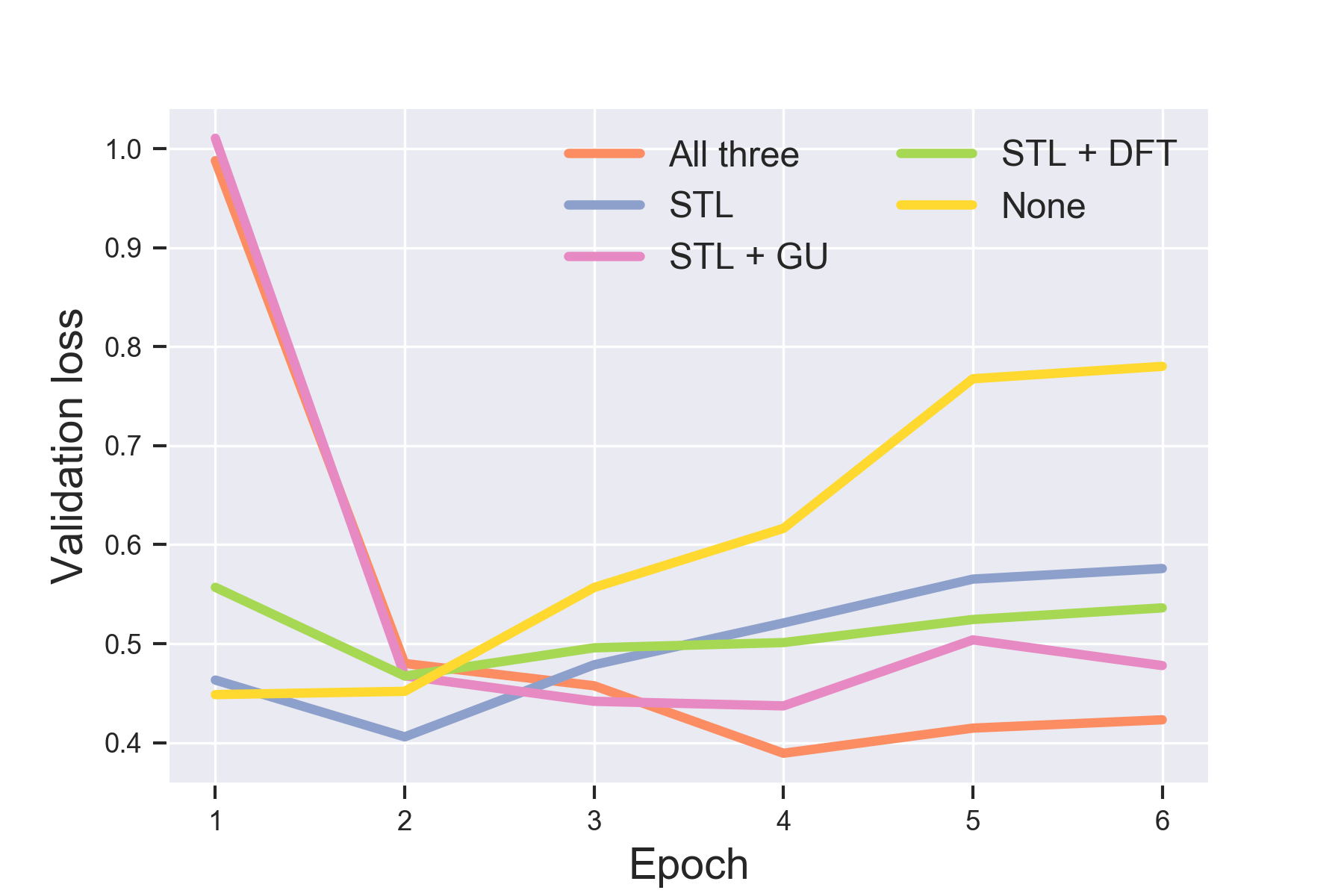}
  \caption{Validation loss trajectories with different training strategies}
  \label{fig:strategy}
  \Description{The 1907 Franklin Model D roadster.}
\end{figure}

For measuring the performance of the techniques against catastrophic forgetting, we try four different settings: No adjustment (NA), only with slanted triangular learning rate (STL), slanted triangular learning rate and gradual unfreezing (STL+GU) and the techniques in the previous one, together with discriminative fine-tuning. We report the performance of these four settings with loss on test function and trajectory of validation loss over training epochs. The results can be seen on table \ref{tab:catastrophic forgetting} and figure \ref{fig:strategy}.

\subfile{tables/catastrophic.tex}

Applying all three of the strategies produce the best performance in terms of test loss and accuracy. Gradual unfreezing and discriminative fine-tuning have the same reasoning behind them: higher level features should be fine-tuned more than the lower level ones, since information learned from language modeling are mostly present in the lower levels. We see from table \ref{tab:catastrophic forgetting} that using only discriminative fine-tuning with slanted triangular learning rates performs worse than using the slanted triangular learning rates alone. This shows that gradual unfreezing is the most important technique for our case.

One way that catastrophic forgetting can show itself is the sudden increase in validation loss after several epochs. As model is trained, it quickly starts to overfit when no measure is taken accordingly. As it can be seen on the figure \ref{fig:strategy}, that is the case when none of the aforementioned techniques are applied. The model achieves the best performance on validation set after the first epoch and then starts to overfit. While with all three techniques applied, model is much more stable. The other combinations lie between these two cases. 

\subsection{Choosing the best layer for classification (RQ5)}
BERT has 12 Transformer encoder layers. It is not necessarily a given that the last layer captures the most relevant information regarding classification task during language model training. For this experiment, we investigate which layer out of 12 Transformer encoder layers give the best result for classification. We put the classification layer after the \textsc{CLS]} tokens of respective representations. We also try taking the average of all layers.

As shown in table \ref{tab:layer4clas}the last layer contributes the most to the model performance in terms of all the metrics measured. This might be indicative of two factors: 1) When the higher layers are used the model that is being trained is larger, hence possibly more powerful, 2) The lower layers capture deeper semantic information, hence they struggle to fine-tune that information for classification.

\subfile{tables/layer4clas.tex}

\subsection{Training only a subset of the layers (RQ6)}
BERT is a very large model. Even on small datasets, fine-tuning the whole model requires significant time and computing power. Therefore if a slightly lower performance can be achieved with fine-tuning only a subset of all parameters, it might be preferable in some contexts. Especially if training set is very large, this change might make BERT more convenient to use. Here we experiment with fine-tuning only the last \textit{k} many encoder layers.

\subfile{tables/unfreezed.tex}

The results are presented on table \ref{tab:unfreezed}. Fine-tuning only the classification layer does not achieve close performance to fine-tuning other layers. However fine-tuning only the last layer handily outperforms the state-of-the-art machine learning methods like HSC. After Layer-9, the performance becomes virtually the same, only to be outperformed by fine-tuning the whole model. This result shows that in order to utilize BERT, an expensive training of the whole model is not mandatory. A fair trade-off can be made for much less training time with a small decrease in model performance. 

\subsection{Where does the model fail?}
With 97\% accuracy on the subset of Financial PhraseBank with 100\% annotator agreement, we think it might be an interesting exercise to examine cases where the model failed to predict the true label. Therefore in this section we will present several examples where model makes the wrong prediction. Also in Malo et al. (2014 )\cite{Malo2014}, it is indicated that most of the inter-annotator disagreements are between positive and neutral labels (agreement for separating positive-negative, negative-neutral and positive-neutral are 98.7\%, 94.2\% and 75.2\% respectively). Authors attribute that the difficulty of distinguishing "commonly used company glitter and actual positive statements". We will present the confusion matrix in order to observe whether this is the case for FinBERT as well. 
\newline

\textbf{Example 1:} \texttt{Pre-tax loss totaled euro 0.3 million , compared to a loss of euro 2.2 million in the first quarter of 2005 .
}

\textbf{True value:} Positive \textbf{Predicted:} Negative
\newline

\textbf{Example 2:} \texttt{This implementation is very important to the operator , since it is about to launch its Fixed to Mobile convergence service in Brazil}

\textbf{True value:} Neutral \textbf{Predicted:} Positive
\newline

\textbf{Example 3:} \texttt{The situation of coated magazine printing paper will continue to be weak .}

\textbf{True value:} Negative \textbf{Predicted:} Neutral
\newline

The first example is actually the most common type of failure. The model fails to do the math in which figure is higher, and in the absence of words indicative of direction like "increased", might make the prediction of neutral. However, there are many similar cases where it does make the true prediction too. Examples 2 and 3 are different versions of the same type of failure. The model fails to distinguish a neutral statement about a given situation from a statement that indicated polarity about the company. In the third example, information about the company's business would probably help.

\begin{figure}[h] 
  \centering
  \includegraphics[width=.6\linewidth]{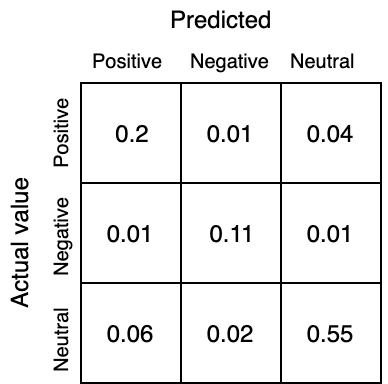}
  \caption{Confusion matrix}
  \label{fig:cm}
  \Description{The 1907 Franklin Model D roadster.}
\end{figure}

The confusion matrix is presented on figure \ref{fig:cm}. 73\% of the failures happen between labels positive and negative, while same number is 5\% for negative and positive. That is consistent with both the inter-annotator agreement numbers and common sense. It is easier to differentiate between positive and negative. But it might be more challenging to decide whether a statement indicates a positive outlook or merely an objective observation.

%% file: tables/finetune_results.tex
\begin{table}
\begin{threeparttable}
  \caption{Performance with different pre-training strategies}
  \label{tab:pretraining}
  \begin{tabular}{lccc}
    \toprule
    Model  & Loss &  Accuracy &  F1 Score\\
    \midrule
    Vanilla BERT &      0.38 & 0.85 & 0.84 \\
    FinBERT-task &      0.39 & 0.86 & \textbf{0.85} \\
    FinBERT-domain   &  \textbf{0.37} & \textbf{0.86} & 0.84 \\
  \bottomrule
\end{tabular}
    \begin{tablenotes}
        \small
        \item \textbf{Bold face} indicates best result in the corresponding metric. Results are reported on 10-fold cross validation.
    \end{tablenotes}

\end{threeparttable}
\end{table}

%% file: tables/catastrophic.tex
\begin{table}
\begin{threeparttable}
  \caption{Performance with different fine-tuning strategies}
  \label{tab:catastrophic forgetting}
  \begin{tabular}{lccc}
    \toprule
    Strategy  &  Loss &  Accuracy &  F1 Score\\
    \midrule
    None  &      0.48 & 0.83 & 0.83 \\
    STL   &       0.40 & 0.81 & 0.82 \\
    STL + GU &  0.40 & 0.86 & \textbf{0.86} \\
    STL + DFT & 0.42 & 0.79 & 0.79 \\
    All three & \textbf{0.37} & \textbf{0.86} & 0.84 \\
  \bottomrule
\end{tabular}

    \begin{tablenotes}
        \small
        \item \textbf{Bold face} indicates best result in the corresponding metric. Results are reported on 10-fold cross validation. STL: slanted triangular learning rates, GU: gradual unfreezing, DFT: discriminative fine-tuning.
    \end{tablenotes}
\end{threeparttable}
\end{table}

%% file: tables/layer4clas.tex
\begin{table}
  \caption{Performance on different encoder layers used for classification}
  \label{tab:layer4clas}
  \begin{tabular}{lccc}
    \toprule
    Layer for classification  &  Loss &  Accuracy &  F1 Score\\
    \midrule
    Layer-1 &      0.65 & 0.76 & 0.77 \\
    Layer-2 &      0.54 & 0.78 & 0.78 \\
    Layer-3 &      0.52 & 0.76 & 0.77 \\
    Layer-4 &      0.48 & 0.80 & 0.77 \\
    Layer-5 &      0.52 & 0.80 & 0.80 \\
    Layer-6 &      0.45 & 0.82 & 0.82 \\
    Layer-7 &      0.43 & 0.82 & 0.83 \\
    Layer-8 &      0.44 & 0.83 & 0.81 \\
    Layer-9 &      0.41 & 0.84 & 0.82 \\
    Layer-10 &      0.42 & 0.83 & 0.82 \\
    Layer-11 &      0.38 & 0.84 & 0.83 \\
    Layer-12 &      0.37 & 0.86 & 0.84 \\
    \midrule
    All layers - mean  &      0.41 & 0.84 & 0.84 \\
  \bottomrule
\end{tabular}
\end{table}

%% file: tables/unfreezed.tex
\begin{table}
  \caption{Performance on starting training from different layers}
  \label{tab:unfreezed}
  \begin{tabular}{lccc}
    \toprule
    First layer unfreezed  &  Loss &  Accuracy &  Training time\\
    \midrule
    Embeddings layer & 0.37 & 0.86 & 332s \\
    Layer-1 &      0.39 & 0.83 & 302s \\
    Layer-2 &      0.39 & 0.83 & 291s \\
    Layer-3 &      0.38 & 0.83 & 272s \\
    Layer-4 &      0.38 & 0.82 & 250s \\
    Layer-5 &      0.40 & 0.83 & 240s \\
    Layer-6 &      0.40 & 0.81 & 220s \\
    Layer-7 &      0.39 & 0.82 & 205s \\
    Layer-8 &      0.39 & 0.84 & 188s \\
    Layer-9 &      0.39 & 0.84 & 172s \\
    Layer-10 &      0.41 & 0.84 & 158s \\
    Layer-11 &      0.45 & 0.82 & 144s \\
    Layer-12 &      0.47 & 0.81 & 133s \\
    Classification layer & 1.04 & 0.52 & 119s \\
  \bottomrule
\end{tabular}
\end{table}

%% file: text/conclusion.tex
\section{Conclusion and Future Work} \label{conclusion}

In this paper, we implemented BERT for the financial domain by further pre-training it on a financial corpus and fine-tuning it for sentiment analysis (FinBERT). This work is the first application of BERT for finance to the best of our knowledge and one of the few that experimented with further pre-training on a domain-specific corpus. On both of the datasets we used, we achieved state-of-the-art results by a significant margin. For the classification task, we increased the state-of-the art by 15\% in accuracy.

In addition to BERT, we also implemented other pre-training language models like ELMo and ULMFit for comparison purposes. ULMFit, further pre-trained on a financial corpus, beat the previous state-of-the art for the classification task, only to a smaller degree than BERT. These results show the effectiveness of pre-trained language models for a down-stream task such as sentiment analysis especially with a small labeled dataset. The complete dataset included more than 3000 examples, but FinBERT was able to surpass the previous state-of-the art even with a training set as small as 500 examples. This is an important result, since deep learning techniques for NLP have been traditionally labeled as too "data-hungry", which is apparently no longer the case.

We conducted extensive experiments with BERT, investigating the effects of further pre-training and several training strategies. We couldn't conclude that further pre-training on a domain-specific corpus was significantly better than not doing so for our case. Our theory is that BERT already performs good enough with our dataset that there is not much room for improvement that further pre-training can provide. We also found that learning rate regimes that fine-tune the higher layers more aggressively than the lower ones perform better and are more effective in preventing catastrophic forgetting. Another conclusion from our experiments was that, comparable performance can be achieved with much less training time by fine-tuning only the last 2 layers of BERT. 

Financial sentiment analysis is not a goal on its own, it is as useful as it can support financial decisions. One way that our work might be extended, could be using FinBERT directly with stock market return data (both in terms of directionality and volatility) on financial news. FinBERT is good enough for extracting explicit sentiments, but modeling implicit information that is not necessarily apparent even to those who are writing the text should be a challenging task. Another possible extension can be using FinBERT for other natural language processing tasks such as named entity recognition or question answering in financial domain. 

\section{Acknowledgements}
I would like to show my gratitude to Pengjie Ren and Zulkuf Genc for their excellent supervision. They provided me with both independence in setting my own course for the research and valuable suggestions when I need them. I would also like to thank Naspers AI team, for entrusting me with this project and always encouraging me to share my work. I am grateful to NIST, for sharing Reuters TRC-2 corpus with me and to Malo et al. for making the excellent Financial PhraseBank publicly available.

%% file: main.bbl

\begin{thebibliography}{33}

%%% the Web address.

\ifx \showCODEN    \undefined \def \showCODEN     #1{\unskip}     \fi
\ifx \showDOI      \undefined \def \showDOI       #1{#1}\fi
\ifx \showISBNx    \undefined \def \showISBNx     #1{\unskip}     \fi
\ifx \showISBNxiii \undefined \def \showISBNxiii  #1{\unskip}     \fi
\ifx \showISSN     \undefined \def \showISSN      #1{\unskip}     \fi
\ifx \showLCCN     \undefined \def \showLCCN      #1{\unskip}     \fi
\ifx \shownote     \undefined \def \shownote      #1{#1}          \fi
\ifx \showarticletitle \undefined \def \showarticletitle #1{#1}   \fi
\ifx \showURL      \undefined \def \showURL       {\relax}        \fi
% invisible to TeX
\providecommand\bibfield[2]{#2}
\providecommand\bibinfo[2]{#2}
\providecommand\natexlab[1]{#1}
\providecommand\showeprint[2][]{arXiv:#2}

\bibitem[\protect\citeauthoryear{Agarwal and Mittal}{Agarwal and
  Mittal}{2016}]%
        {Agarwal2016}
\bibfield{author}{\bibinfo{person}{Basant Agarwal} {and}
  \bibinfo{person}{Namita Mittal}.} \bibinfo{year}{2016}\natexlab{}.
\newblock \bibinfo{booktitle}{\emph{Machine Learning Approach for Sentiment
  Analysis}}.
\newblock \bibinfo{publisher}{Springer International Publishing},
  \bibinfo{address}{Cham}, \bibinfo{pages}{21--45}.
\newblock
\showISBNx{978-3-319-25343-5}
\urldef\tempurl%
\url{https://doi.org/10.1007/978-3-319-25343-5_3}
\showDOI{\tempurl}


\bibitem[\protect\citeauthoryear{Araque, Corcuera-Platas, S{\'{a}}nchez-Rada,
  and Iglesias}{Araque et~al\mbox{.}}{2017}]%
        {Araque2017}
\bibfield{author}{\bibinfo{person}{Oscar Araque}, \bibinfo{person}{Ignacio
  Corcuera-Platas}, \bibinfo{person}{J.~Fernando S{\'{a}}nchez-Rada}, {and}
  \bibinfo{person}{Carlos~A. Iglesias}.} \bibinfo{year}{2017}\natexlab{}.
\newblock \showarticletitle{Enhancing deep learning sentiment analysis with
  ensemble techniques in social applications}.
\newblock \bibinfo{journal}{\emph{Expert Systems with Applications}}
  \bibinfo{volume}{77} (\bibinfo{date}{jul} \bibinfo{year}{2017}),
  \bibinfo{pages}{236--246}.
\newblock
\urldef\tempurl%
\url{https://doi.org/10.1016/j.eswa.2017.02.002}
\showDOI{\tempurl}


\bibitem[\protect\citeauthoryear{Devlin, Chang, Lee, and Toutanova}{Devlin
  et~al\mbox{.}}{2018}]%
        {Devlin2018}
\bibfield{author}{\bibinfo{person}{Jacob Devlin}, \bibinfo{person}{Ming-Wei
  Chang}, \bibinfo{person}{Kenton Lee}, {and} \bibinfo{person}{Kristina
  Toutanova}.} \bibinfo{year}{2018}\natexlab{}.
\newblock \showarticletitle{{BERT: Pre-training of Deep Bidirectional
  Transformers for Language Understanding}}.
\newblock  (\bibinfo{year}{2018}).
\newblock
\showISBNx{0-674-24915-1 0-674-24917-8}
\showISSN{0140-525X}
\urldef\tempurl%
\url{https://doi.org/arXiv:1811.03600v2}
\showDOI{\tempurl}
\showeprint[arxiv]{1810.04805}


\bibitem[\protect\citeauthoryear{Guo, Shi, and Tu}{Guo et~al\mbox{.}}{2016}]%
        {Guo2016}
\bibfield{author}{\bibinfo{person}{Li Guo}, \bibinfo{person}{Feng Shi}, {and}
  \bibinfo{person}{Jun Tu}.} \bibinfo{year}{2016}\natexlab{}.
\newblock \showarticletitle{{Textual analysis and machine leaning: Crack
  unstructured data in finance and accounting}}.
\newblock \bibinfo{journal}{\emph{The Journal of Finance and Data Science}}
  \bibinfo{volume}{2}, \bibinfo{number}{3} (\bibinfo{date}{sep}
  \bibinfo{year}{2016}), \bibinfo{pages}{153--170}.
\newblock
\showISSN{2405-9188}
\urldef\tempurl%
\url{https://doi.org/10.1016/J.JFDS.2017.02.001}
\showDOI{\tempurl}


\bibitem[\protect\citeauthoryear{Howard and Ruder}{Howard and Ruder}{2018}]%
        {Howard2018}
\bibfield{author}{\bibinfo{person}{Jeremy Howard} {and}
  \bibinfo{person}{Sebastian Ruder}.} \bibinfo{year}{2018}\natexlab{}.
\newblock \showarticletitle{{Universal Language Model Fine-tuning for Text
  Classification}}.
\newblock  (\bibinfo{date}{jan} \bibinfo{year}{2018}).
\newblock
\showeprint[arxiv]{1801.06146}
\urldef\tempurl%
\url{http://arxiv.org/abs/1801.06146}
\showURL{%
\tempurl}


\bibitem[\protect\citeauthoryear{Kant, Puri, Yakovenko, and Catanzaro}{Kant
  et~al\mbox{.}}{2018}]%
        {Kant2018}
\bibfield{author}{\bibinfo{person}{Neel Kant}, \bibinfo{person}{Raul Puri},
  \bibinfo{person}{Nikolai Yakovenko}, {and} \bibinfo{person}{Bryan
  Catanzaro}.} \bibinfo{year}{2018}\natexlab{}.
\newblock \showarticletitle{{Practical Text Classification With Large
  Pre-Trained Language Models}}.
\newblock  (\bibinfo{year}{2018}).
\newblock
\showISBNx{1812.01207v1}
\showeprint[arxiv]{1812.01207}
\urldef\tempurl%
\url{http://arxiv.org/abs/1812.01207}
\showURL{%
\tempurl}


\bibitem[\protect\citeauthoryear{Kraus and Feuerriegel}{Kraus and
  Feuerriegel}{2017}]%
        {Kraus2017}
\bibfield{author}{\bibinfo{person}{Mathias Kraus} {and} \bibinfo{person}{Stefan
  Feuerriegel}.} \bibinfo{year}{2017}\natexlab{}.
\newblock \showarticletitle{{Decision support from financial disclosures with
  deep neural networks and transfer learning}}.
\newblock \bibinfo{journal}{\emph{Decision Support Systems}}
  \bibinfo{volume}{104} (\bibinfo{year}{2017}), \bibinfo{pages}{38--48}.
\newblock
\showISBNx{1664-1078}
\showISSN{01679236}
\urldef\tempurl%
\url{https://doi.org/10.1016/j.dss.2017.10.001}
\showDOI{\tempurl}
\showeprint[arxiv]{1710.03954}


\bibitem[\protect\citeauthoryear{Krishnamoorthy}{Krishnamoorthy}{2018}]%
        {Krishnamoorthy2018}
\bibfield{author}{\bibinfo{person}{Srikumar Krishnamoorthy}.}
  \bibinfo{year}{2018}\natexlab{}.
\newblock \showarticletitle{{Sentiment analysis of financial news articles
  using performance indicators}}.
\newblock \bibinfo{journal}{\emph{Knowledge and Information Systems}}
  \bibinfo{volume}{56}, \bibinfo{number}{2} (\bibinfo{date}{aug}
  \bibinfo{year}{2018}), \bibinfo{pages}{373--394}.
\newblock
\showISSN{02193116}
\urldef\tempurl%
\url{https://doi.org/10.1007/s10115-017-1134-1}
\showDOI{\tempurl}


\bibitem[\protect\citeauthoryear{Li, Xie, Chen, Wang, and Deng}{Li
  et~al\mbox{.}}{2014}]%
        {Li2014}
\bibfield{author}{\bibinfo{person}{Xiaodong Li}, \bibinfo{person}{Haoran Xie},
  \bibinfo{person}{Li Chen}, \bibinfo{person}{Jianping Wang}, {and}
  \bibinfo{person}{Xiaotie Deng}.} \bibinfo{year}{2014}\natexlab{}.
\newblock \showarticletitle{News impact on stock price return via sentiment
  analysis}.
\newblock \bibinfo{journal}{\emph{Knowledge-Based Systems}}
  \bibinfo{volume}{69} (\bibinfo{date}{oct} \bibinfo{year}{2014}),
  \bibinfo{pages}{14--23}.
\newblock
\urldef\tempurl%
\url{https://doi.org/10.1016/j.knosys.2014.04.022}
\showDOI{\tempurl}


\bibitem[\protect\citeauthoryear{Liu}{Liu}{2012}]%
        {Liu2012}
\bibfield{author}{\bibinfo{person}{Bing Liu}.} \bibinfo{year}{2012}\natexlab{}.
\newblock \showarticletitle{Sentiment Analysis and Opinion Mining}.
\newblock \bibinfo{journal}{\emph{Synthesis Lectures on Human Language
  Technologies}} \bibinfo{volume}{5}, \bibinfo{number}{1} (\bibinfo{date}{may}
  \bibinfo{year}{2012}), \bibinfo{pages}{1--167}.
\newblock
\urldef\tempurl%
\url{https://doi.org/10.2200/s00416ed1v01y201204hlt016}
\showDOI{\tempurl}


\bibitem[\protect\citeauthoryear{Loughran and Mcdonald}{Loughran and
  Mcdonald}{2011}]%
        {Loughran2011}
\bibfield{author}{\bibinfo{person}{Tim Loughran} {and} \bibinfo{person}{Bill
  Mcdonald}.} \bibinfo{year}{2011}\natexlab{}.
\newblock \showarticletitle{{When Is a Liability Not a Liability? Textual
  Analysis, Dictionaries, and 10-Ks}}.
\newblock \bibinfo{journal}{\emph{Journal of Finance}} \bibinfo{volume}{66},
  \bibinfo{number}{1} (\bibinfo{date}{feb} \bibinfo{year}{2011}),
  \bibinfo{pages}{35--65}.
\newblock
\showISSN{00221082}
\urldef\tempurl%
\url{https://doi.org/10.1111/j.1540-6261.2010.01625.x}
\showDOI{\tempurl}


\bibitem[\protect\citeauthoryear{Loughran and Mcdonald}{Loughran and
  Mcdonald}{2016}]%
        {Loughran2016}
\bibfield{author}{\bibinfo{person}{Tim Loughran} {and} \bibinfo{person}{Bill
  Mcdonald}.} \bibinfo{year}{2016}\natexlab{}.
\newblock \showarticletitle{{Textual Analysis in Accounting and Finance: A
  Survey}}.
\newblock \bibinfo{journal}{\emph{Journal of Accounting Research}}
  \bibinfo{volume}{54}, \bibinfo{number}{4} (\bibinfo{year}{2016}),
  \bibinfo{pages}{1187--1230}.
\newblock
\showISBNx{ID 2504147}
\showISSN{1475679X}
\urldef\tempurl%
\url{https://doi.org/10.1111/1475-679X.12123}
\showDOI{\tempurl}


\bibitem[\protect\citeauthoryear{Lutz, Pr{\"{o}}llochs, and Neumann}{Lutz
  et~al\mbox{.}}{2018}]%
        {Lutz2018}
\bibfield{author}{\bibinfo{person}{Bernhard Lutz}, \bibinfo{person}{Nicolas
  Pr{\"{o}}llochs}, {and} \bibinfo{person}{Dirk Neumann}.}
  \bibinfo{year}{2018}\natexlab{}.
\newblock \bibinfo{booktitle}{\emph{{Sentence-Level Sentiment Analysis of
  Financial News Using Distributed Text Representations and Multi-Instance
  Learning}}}.
\newblock \bibinfo{type}{{T}echnical {R}eport}.
\newblock
\showeprint[arxiv]{1901.00400}
\urldef\tempurl%
\url{http://arxiv.org/abs/1901.00400}
\showURL{%
\tempurl}


\bibitem[\protect\citeauthoryear{Maia, Freitas, and Handschuh}{Maia
  et~al\mbox{.}}{2018a}]%
        {Maia20182}
\bibfield{author}{\bibinfo{person}{Macedo Maia}, \bibinfo{person}{Andr�
  Freitas}, {and} \bibinfo{person}{Siegfried Handschuh}.}
  \bibinfo{year}{2018}\natexlab{a}.
\newblock \showarticletitle{{FinSSLx: A Sentiment Analysis Model for the
  Financial Domain Using Text Simplification}}. In
  \bibinfo{booktitle}{\emph{2018 IEEE 12th International Conference on Semantic
  Computing (ICSC)}}. \bibinfo{publisher}{IEEE}, \bibinfo{pages}{318--319}.
\newblock
\showISBNx{978-1-5386-4408-9}
\urldef\tempurl%
\url{https://doi.org/10.1109/ICSC.2018.00065}
\showDOI{\tempurl}


\bibitem[\protect\citeauthoryear{Maia, Handschuh, Freitas, Davis, Mcdermott,
  Zarrouk, Balahur, and Mc-Dermott}{Maia et~al\mbox{.}}{2018b}]%
        {Maia2018}
\bibfield{author}{\bibinfo{person}{Macedo Maia}, \bibinfo{person}{Siegfried
  Handschuh}, \bibinfo{person}{Andr{\'{e}} Freitas}, \bibinfo{person}{Brian
  Davis}, \bibinfo{person}{Ross Mcdermott}, \bibinfo{person}{Manel Zarrouk},
  \bibinfo{person}{Alexandra Balahur}, {and} \bibinfo{person}{Ross
  Mc-Dermott}.} \bibinfo{year}{2018}\natexlab{b}.
\newblock \showarticletitle{{Companion of the The Web Conference 2018 on The
  Web Conference 2018, {\{}WWW{\}} 2018, Lyon , France, April 23-27, 2018}}.
  \bibinfo{publisher}{ACM}.
\newblock
\showISBNx{9781450356404}
\urldef\tempurl%
\url{https://doi.org/10.1145/3184558}
\showDOI{\tempurl}


\bibitem[\protect\citeauthoryear{Malkiel}{Malkiel}{2003}]%
        {Malkiel2003}
\bibfield{author}{\bibinfo{person}{Burton~G Malkiel}.}
  \bibinfo{year}{2003}\natexlab{}.
\newblock \showarticletitle{{The Efficient Market Hypothesis and Its Critics}}.
\newblock \bibinfo{journal}{\emph{Journal of Economic Perspectives}}
  \bibinfo{volume}{17}, \bibinfo{number}{1} (\bibinfo{date}{feb}
  \bibinfo{year}{2003}), \bibinfo{pages}{59--82}.
\newblock
\showISSN{0895-3309}
\urldef\tempurl%
\url{https://doi.org/10.1257/089533003321164958}
\showDOI{\tempurl}


\bibitem[\protect\citeauthoryear{Malo, Sinha, Korhonen, Wallenius, and
  Takala}{Malo et~al\mbox{.}}{2014}]%
        {Malo2014}
\bibfield{author}{\bibinfo{person}{Pekka Malo}, \bibinfo{person}{Ankur Sinha},
  \bibinfo{person}{Pekka Korhonen}, \bibinfo{person}{Jyrki Wallenius}, {and}
  \bibinfo{person}{Pyry Takala}.} \bibinfo{year}{2014}\natexlab{}.
\newblock \showarticletitle{{Good debt or bad debt: Detecting semantic
  orientations in economic texts}}.
\newblock \bibinfo{journal}{\emph{Journal of the Association for Information
  Science and Technology}} \bibinfo{volume}{65}, \bibinfo{number}{4}
  (\bibinfo{year}{2014}), \bibinfo{pages}{782--796}.
\newblock
\showISSN{23301643}
\urldef\tempurl%
\url{https://doi.org/10.1002/asi.23062}
\showDOI{\tempurl}
\showeprint[arxiv]{arXiv:1307.5336v2}


\bibitem[\protect\citeauthoryear{{Marcus}}{{Marcus}}{2018}]%
        {2018arXiv180100631M}
\bibfield{author}{\bibinfo{person}{G. {Marcus}}.}
  \bibinfo{year}{2018}\natexlab{}.
\newblock \showarticletitle{{Deep Learning: A Critical Appraisal}}.
\newblock \bibinfo{journal}{\emph{arXiv e-prints}} (\bibinfo{date}{Jan.}
  \bibinfo{year}{2018}).
\newblock
\showeprint[arxiv]{cs.AI/1801.00631}


\bibitem[\protect\citeauthoryear{Martineau and Finin}{Martineau and
  Finin}{2009}]%
        {conf/icwsm/MartineauF09}
\bibfield{author}{\bibinfo{person}{Justin Martineau} {and} \bibinfo{person}{Tim
  Finin}.} \bibinfo{year}{2009}\natexlab{}.
\newblock \showarticletitle{Delta TFIDF: An Improved Feature Space for
  Sentiment Analysis.}. In \bibinfo{booktitle}{\emph{ICWSM}},
  \bibfield{editor}{\bibinfo{person}{Eytan Adar}, \bibinfo{person}{Matthew
  Hurst}, \bibinfo{person}{Tim Finin}, \bibinfo{person}{Natalie~S. Glance},
  \bibinfo{person}{Nicolas Nicolov}, {and} \bibinfo{person}{Belle~L. Tseng}}
  (Eds.). \bibinfo{publisher}{The AAAI Press}.
\newblock
\showISBNx{978-1-57735-421-5}
\urldef\tempurl%
\url{http://dblp.uni-trier.de/db/conf/icwsm/icwsm2009.html#MartineauF09}
\showURL{%
\tempurl}


\bibitem[\protect\citeauthoryear{McCann, Bradbury, Xiong, and Socher}{McCann
  et~al\mbox{.}}{2017}]%
        {McCann2017}
\bibfield{author}{\bibinfo{person}{Bryan McCann}, \bibinfo{person}{James
  Bradbury}, \bibinfo{person}{Caiming Xiong}, {and} \bibinfo{person}{Richard
  Socher}.} \bibinfo{year}{2017}\natexlab{}.
\newblock \showarticletitle{{Learned in Translation: Contextualized Word
  Vectors}}.
\newblock  \bibinfo{number}{Nips} (\bibinfo{year}{2017}),
  \bibinfo{pages}{1--12}.
\newblock
\showeprint[arxiv]{1708.00107}
\urldef\tempurl%
\url{http://arxiv.org/abs/1708.00107}
\showURL{%
\tempurl}


\bibitem[\protect\citeauthoryear{Merity, Keskar, and Socher}{Merity
  et~al\mbox{.}}{2017}]%
        {DBLP:journals/corr/abs-1708-02182}
\bibfield{author}{\bibinfo{person}{Stephen Merity},
  \bibinfo{person}{Nitish~Shirish Keskar}, {and} \bibinfo{person}{Richard
  Socher}.} \bibinfo{year}{2017}\natexlab{}.
\newblock \showarticletitle{Regularizing and Optimizing {LSTM} Language
  Models}.
\newblock \bibinfo{journal}{\emph{CoRR}}  \bibinfo{volume}{abs/1708.02182}
  (\bibinfo{year}{2017}).
\newblock
\showeprint[arxiv]{1708.02182}
\urldef\tempurl%
\url{http://arxiv.org/abs/1708.02182}
\showURL{%
\tempurl}


\bibitem[\protect\citeauthoryear{Pennington, Socher, and Manning}{Pennington
  et~al\mbox{.}}{2014}]%
        {pennington-etal-2014-glove}
\bibfield{author}{\bibinfo{person}{Jeffrey Pennington},
  \bibinfo{person}{Richard Socher}, {and} \bibinfo{person}{Christopher
  Manning}.} \bibinfo{year}{2014}\natexlab{}.
\newblock \showarticletitle{{G}love: Global Vectors for Word Representation}.
  In \bibinfo{booktitle}{\emph{Proceedings of the 2014 Conference on Empirical
  Methods in Natural Language Processing ({EMNLP})}}.
  \bibinfo{publisher}{Association for Computational Linguistics},
  \bibinfo{address}{Doha, Qatar}, \bibinfo{pages}{1532--1543}.
\newblock
\urldef\tempurl%
\url{https://doi.org/10.3115/v1/D14-1162}
\showDOI{\tempurl}


\bibitem[\protect\citeauthoryear{Peters, Neumann, Iyyer, Gardner, Clark, Lee,
  and Zettlemoyer}{Peters et~al\mbox{.}}{2018}]%
        {Peters2018}
\bibfield{author}{\bibinfo{person}{Matthew~E Peters}, \bibinfo{person}{Mark
  Neumann}, \bibinfo{person}{Mohit Iyyer}, \bibinfo{person}{Matt Gardner},
  \bibinfo{person}{Christopher Clark}, \bibinfo{person}{Kenton Lee}, {and}
  \bibinfo{person}{Luke Zettlemoyer}.} \bibinfo{year}{2018}\natexlab{}.
\newblock \showarticletitle{{Deep contextualized word representations}}.
\newblock  (\bibinfo{year}{2018}).
\newblock
\showISBNx{9781941643327}
\showISSN{9781941643327}
\urldef\tempurl%
\url{https://doi.org/10.18653/v1/N18-1202}
\showDOI{\tempurl}
\showeprint[arxiv]{1802.05365}


\bibitem[\protect\citeauthoryear{Piao and Breslin}{Piao and Breslin}{2018}]%
        {Piao2018}
\bibfield{author}{\bibinfo{person}{Guangyuan Piao} {and}
  \bibinfo{person}{John~G Breslin}.} \bibinfo{year}{2018}\natexlab{}.
\newblock \showarticletitle{{Financial Aspect and Sentiment Predictions with
  Deep Neural Networks}}. \bibinfo{pages}{1973--1977}.
\newblock
\showISBNx{9781450356404}
\urldef\tempurl%
\url{https://doi.org/10.1145/3184558.3191829}
\showDOI{\tempurl}


\bibitem[\protect\citeauthoryear{Severyn and Moschitti}{Severyn and
  Moschitti}{2015}]%
        {Severyn2015}
\bibfield{author}{\bibinfo{person}{Aliaksei Severyn} {and}
  \bibinfo{person}{Alessandro Moschitti}.} \bibinfo{year}{2015}\natexlab{}.
\newblock \showarticletitle{Twitter Sentiment Analysis with Deep Convolutional
  Neural Networks}. In \bibinfo{booktitle}{\emph{Proceedings of the 38th
  International {ACM} {SIGIR} Conference on Research and Development in
  Information Retrieval - {SIGIR} {\textquotesingle}15}}.
  \bibinfo{publisher}{{ACM} Press}.
\newblock
\urldef\tempurl%
\url{https://doi.org/10.1145/2766462.2767830}
\showDOI{\tempurl}


\bibitem[\protect\citeauthoryear{Sohangir, Wang, Pomeranets, and
  Khoshgoftaar}{Sohangir et~al\mbox{.}}{2018}]%
        {Sohangir2018}
\bibfield{author}{\bibinfo{person}{Sahar Sohangir}, \bibinfo{person}{Dingding
  Wang}, \bibinfo{person}{Anna Pomeranets}, {and} \bibinfo{person}{Taghi~M
  Khoshgoftaar}.} \bibinfo{year}{2018}\natexlab{}.
\newblock \showarticletitle{{Big Data: Deep Learning for financial sentiment
  analysis}}.
\newblock \bibinfo{journal}{\emph{Journal of Big Data}} \bibinfo{volume}{5},
  \bibinfo{number}{1} (\bibinfo{year}{2018}).
\newblock
\showISBNx{4053701701116}
\showISSN{21961115}
\urldef\tempurl%
\url{https://doi.org/10.1186/s40537-017-0111-6}
\showDOI{\tempurl}


\bibitem[\protect\citeauthoryear{Sun, Qiu, Xu, and Huang}{Sun
  et~al\mbox{.}}{2019}]%
        {Sun2019}
\bibfield{author}{\bibinfo{person}{Chi Sun}, \bibinfo{person}{Xipeng Qiu},
  \bibinfo{person}{Yige Xu}, {and} \bibinfo{person}{Xuanjing Huang}.}
  \bibinfo{year}{2019}\natexlab{}.
\newblock \showarticletitle{{How to Fine-Tune BERT for Text Classification?}}
\newblock  (\bibinfo{year}{2019}).
\newblock
\showeprint[arxiv]{1905.05583}
\urldef\tempurl%
\url{https://arxiv.org/pdf/1905.05583v1.pdf http://arxiv.org/abs/1905.05583}
\showURL{%
\tempurl}


\bibitem[\protect\citeauthoryear{Tripathy, Agrawal, and Rath}{Tripathy
  et~al\mbox{.}}{2016}]%
        {Tripathy2016}
\bibfield{author}{\bibinfo{person}{Abinash Tripathy}, \bibinfo{person}{Ankit
  Agrawal}, {and} \bibinfo{person}{Santanu~Kumar Rath}.}
  \bibinfo{year}{2016}\natexlab{}.
\newblock \showarticletitle{Classification of sentiment reviews using n-gram
  machine learning approach}.
\newblock \bibinfo{journal}{\emph{Expert Systems with Applications}}
  \bibinfo{volume}{57} (\bibinfo{date}{sep} \bibinfo{year}{2016}),
  \bibinfo{pages}{117--126}.
\newblock
\urldef\tempurl%
\url{https://doi.org/10.1016/j.eswa.2016.03.028}
\showDOI{\tempurl}


\bibitem[\protect\citeauthoryear{Vaswani, Shazeer, Parmar, Uszkoreit, Jones,
  Gomez, Kaiser, and Polosukhin}{Vaswani et~al\mbox{.}}{2017}]%
        {Vaswani2017}
\bibfield{author}{\bibinfo{person}{Ashish Vaswani}, \bibinfo{person}{Noam
  Shazeer}, \bibinfo{person}{Niki Parmar}, \bibinfo{person}{Jakob Uszkoreit},
  \bibinfo{person}{Llion Jones}, \bibinfo{person}{Aidan~N. Gomez},
  \bibinfo{person}{Lukasz Kaiser}, {and} \bibinfo{person}{Illia Polosukhin}.}
  \bibinfo{year}{2017}\natexlab{}.
\newblock \showarticletitle{{Attention Is All You Need}}.
\newblock  \bibinfo{number}{Nips} (\bibinfo{year}{2017}).
\newblock
\showeprint[arxiv]{1706.03762}
\urldef\tempurl%
\url{http://arxiv.org/abs/1706.03762}
\showURL{%
\tempurl}


\bibitem[\protect\citeauthoryear{Whitelaw, Garg, and Argamon}{Whitelaw
  et~al\mbox{.}}{2005}]%
        {Whitelaw2005}
\bibfield{author}{\bibinfo{person}{Casey Whitelaw}, \bibinfo{person}{Navendu
  Garg}, {and} \bibinfo{person}{Shlomo Argamon}.}
  \bibinfo{year}{2005}\natexlab{}.
\newblock \showarticletitle{Using appraisal groups for sentiment analysis}. In
  \bibinfo{booktitle}{\emph{Proceedings of the 14th {ACM} international
  conference on Information and knowledge management - {CIKM}
  {\textquotesingle}05}}. \bibinfo{publisher}{{ACM} Press}.
\newblock
\urldef\tempurl%
\url{https://doi.org/10.1145/1099554.1099714}
\showDOI{\tempurl}


\bibitem[\protect\citeauthoryear{Yang, Rosenfeld, and Makutonin}{Yang
  et~al\mbox{.}}{2018}]%
        {Yang2018}
\bibfield{author}{\bibinfo{person}{Steve Yang}, \bibinfo{person}{Jason
  Rosenfeld}, {and} \bibinfo{person}{Jacques Makutonin}.}
  \bibinfo{year}{2018}\natexlab{}.
\newblock \showarticletitle{{Financial Aspect-Based Sentiment Analysis using
  Deep Representations}}.
\newblock  (\bibinfo{year}{2018}).
\newblock
\showeprint[arxiv]{1808.07931}
\urldef\tempurl%
\url{https://arxiv.org/pdf/1808.07931v1.pdf http://arxiv.org/abs/1808.07931}
\showURL{%
\tempurl}


\bibitem[\protect\citeauthoryear{Zhang, Wang, and Liu}{Zhang
  et~al\mbox{.}}{2018}]%
        {Zhang2018}
\bibfield{author}{\bibinfo{person}{Lei Zhang}, \bibinfo{person}{Shuai Wang},
  {and} \bibinfo{person}{Bing Liu}.} \bibinfo{year}{2018}\natexlab{}.
\newblock \showarticletitle{Deep learning for sentiment analysis: A survey}.
\newblock \bibinfo{journal}{\emph{Wiley Interdisciplinary Reviews: Data Mining
  and Knowledge Discovery}} \bibinfo{volume}{8}, \bibinfo{number}{4}
  (\bibinfo{date}{mar} \bibinfo{year}{2018}), \bibinfo{pages}{e1253}.
\newblock
\urldef\tempurl%
\url{https://doi.org/10.1002/widm.1253}
\showDOI{\tempurl}


\bibitem[\protect\citeauthoryear{Zhu, Kiros, Zemel, Salakhutdinov, Urtasun,
  Torralba, and Fidler}{Zhu et~al\mbox{.}}{2015}]%
        {Zhu2015}
\bibfield{author}{\bibinfo{person}{Yukun Zhu}, \bibinfo{person}{Ryan Kiros},
  \bibinfo{person}{Richard Zemel}, \bibinfo{person}{Ruslan Salakhutdinov},
  \bibinfo{person}{Raquel Urtasun}, \bibinfo{person}{Antonio Torralba}, {and}
  \bibinfo{person}{Sanja Fidler}.} \bibinfo{year}{2015}\natexlab{}.
\newblock \showarticletitle{{Aligning Books and Movies: Towards Story-like
  Visual Explanations by Watching Movies and Reading Books}}.
\newblock  (\bibinfo{date}{jun} \bibinfo{year}{2015}).
\newblock
\showeprint[arxiv]{1506.06724}
\urldef\tempurl%
\url{http://arxiv.org/abs/1506.06724}
\showURL{%
\tempurl}


\end{thebibliography}
